# Deep Learning in Image Classification: Evaluating VGG19's Performance on Complex Visual Data

Weijie He
University of California, Los Angeles
Los Angeles, USA

Tong Zhou
Rice University
Houston, USA

Yanlin Xiang
University of Houston
Houston, USA

Yang Lin
University of Pennsylvania
Philadelphia, USA

Jiacheng Hu
Tulane University
New Orleans, USA

Runyuan Bao*
Johns Hopkins University
Baltimore, USA

*Abstract*—This study aims to explore the automatic classification method of pneumonia X-ray images based on VGG19 deep convolutional neural network, and evaluate its application effect in pneumonia diagnosis by comparing with classic models such as SVM, XGBoost, MLP, and ResNet50. The experimental results show that VGG19 performs well in multiple indicators such as accuracy (92%), AUC (0.95), F1 score (0.90) and recall rate (0.87), which is better than other comparison models, especially in image feature extraction and classification accuracy. Although ResNet50 performs well in some indicators, it is slightly inferior to VGG19 in recall rate and F1 score. Traditional machine learning models SVM and XGBoost are obviously limited in image classification tasks, especially in complex medical image analysis tasks, and their performance is relatively mediocre. The research results show that deep learning, especially convolutional neural networks, have significant advantages in medical image classification tasks, especially in pneumonia X-ray image analysis, and can provide efficient and accurate automatic diagnosis support. This research provides strong technical support for the early detection of pneumonia and the development of automated diagnosis systems and also lays the foundation for further promoting the application and development of automated medical image processing technology.

## I. INTRODUCTION

Pneumonia, as a common respiratory disease, has long caused a large number of health problems worldwide. According to statistics from the World Health Organization, pneumonia remains one of the major causes of death in children and the elderly. Traditional pneumonia diagnosis usually relies on multiple factors such as clinical symptoms, body temperature, cough, and chest X-ray images. However, the interpretation and analysis of X-ray images rely on the experience of radiologists, which is subjective and human, and is inefficient when dealing with large numbers of cases. With the rapid development of medical imaging technology and deep learning technology, pneumonia diagnosis methods based on automated image analysis have gradually become a research hotspot. In particular, the introduction of convolutional neural networks (CNNs) has made it possible to automatically classify and identify pneumonia X-ray images [1], and has provided new ideas and methods for early diagnosis and precise treatment of pneumonia.

Deep convolutional neural networks, specifically VGG19 [2], have achieved very impressive results on many medical image classification tasks. VGG19 is a 19-weight layer deep convolutional neural network architecture. Having a simple structure and being rather easy to train, it has achieved excellent performance in numerous image classification problems. The advantage of VGG19 is that its deep network structure can extract more complex image features, which means, in particular when working with fine-grained classification tasks, VGG19 is able to identify even the smallest differences in pictures. For X-ray images with pneumonia, VGG19 may learn from the image and automatically grasp the local features related to pneumonia, giving more accurate outcomes for the classification [3]. In conclusion, the pneumonia X-ray image classification based on VGG19 has enhanced diagnosis accuracy and greatly reduced time and workload in manual interpretation.

VGG19's success in pneumonia X-ray classification lies in its ability to extract detailed features from complex medical images. Its multi-layer convolutional design enables accurate detection of abnormal lung regions and classification of pneumonia types, making it a reliable tool for clinical diagnosis. Leveraging transfer learning, VGG19 uses pre-trained weights to accelerate training, improve accuracy, and reduce reliance on labeled data. Transfer learning has been extensively validated in diverse domains, including natural language processing (NLP) [4] and financial risk analysis [5], demonstrating its ability to adapt pre-trained models to domain-specific tasks efficiently. In NLP, techniques like BERT and GPT [6] have set benchmarks by transferring linguistic knowledge across tasks [7-9], while in financial risk analysis, pre-trained models have enhanced predictive accuracy for credit scoring [10], fraud detection [11], and market risk assessments [12-13]. These successes underscore the versatility and effectiveness of transfer learning in complex, data-intensive fields. This efficiency allows it to adapt quickly to new tasks, addressing clinical challenges effectively. With advancements in computing and data, VGG19 continues to improve its precision across diverse patient groups and pneumonia types. Its integration into medical systems enhances early diagnosis and treatment, reducing mortality and advancing intelligent healthcare solutions.

## II. RELATED WORK

The advancement of deep learning, particularly convolutional neural networks (CNNs), has significantly impacted the field of medical image classification. This study builds on existing research, leveraging the capabilities of VGG19 for pneumonia X-ray image classification, and draws on prior contributions in transfer learning, data augmentation, and multimodal approaches.

Deep convolutional neural networks have consistently demonstrated their strength in extracting complex features from medical images. Zheng et al. [14] highlighted the capabilities of fully convolutional networks for high-precision medical image analysis, providing a strong foundation for adopting CNN architectures in tasks requiring fine-grained feature extraction. Similarly, Xiao et al. [15] used CNNs to classify breast cancer cytopathology images, emphasizing their adaptability to diverse medical datasets, a principle applied in this study to analyze pneumonia X-rays.

Transfer learning, a critical aspect of this study, has proven effective in improving model performance with limited data. Wang et al. [16] demonstrated how deep transfer learning could enhance breast cancer image classification, showing that pre-trained models significantly reduce training time and improve accuracy. Lu et al. [17] extended this approach through federated learning, enabling scalable vision-and-language representation learning, a technique relevant for improving generalization in complex medical image analysis.

Semi-supervised and self-supervised learning methods have addressed the challenges posed by limited labeled data. Shen et al. [18] proposed leveraging semi-supervised learning to enhance image classification under data scarcity, while Xiao [19] explored self-supervised pathways to improve generalization in few-shot scenarios. These techniques align with this study's reliance on pre-trained models and efficient data utilization to optimize performance. Ruan et al. [20] provided a comprehensive evaluation of multimodal AI models in medical imaging, emphasizing data augmentation and model comparison techniques. Their insights validate the approach taken in this research to compare VGG19 with other architectures, such as ResNet50, while ensuring robustness in classification performance across different evaluation metrics. Broader applications of CNNs in medical imaging have further supported their adoption for pneumonia diagnosis. Yan et al. [21] applied neural networks for survival prediction across diverse cancer types, illustrating the versatility of CNN-based architectures in medical data analysis. Additionally, Sun et al. [22] explored convolutional networks for gesture recognition, highlighting their adaptability to tasks requiring high-dimensional feature extraction, which complements the requirements of pneumonia X-ray classification.

These studies collectively demonstrate the efficacy of CNNs, transfer learning, and advanced data handling methods in addressing complex image classification tasks, forming the basis for applying VGG19 in this study to improve pneumonia diagnosis accuracy.

## III. METHOD

In this study, we used an automatic classification method for pneumonia X-ray images based on the VGG19 deep convolutional neural network (CNN). As a typical convolutional neural network model, VGG19 has been widely used in various image classification tasks. Its structural depth and simple convolutional layer design make it one of the classic models in the field of image processing. In the classification task of pneumonia X-ray images, VGG19 can effectively extract high-dimensional features of images and perform nonlinear mapping through deep neural networks to achieve accurate classification of pneumonia images. Its model architecture is shown in Figure 1.

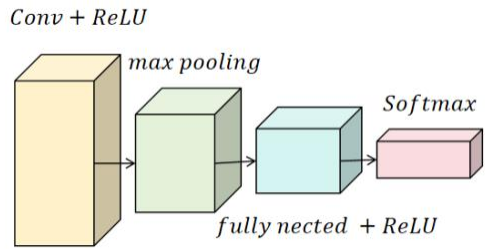

Figure 1 Overall model architecture

The core idea of the VGG19 model is to extract low-level and high-level features of an image by stacking multiple convolutional layers and pooling layers. The main function of the convolutional layer is to extract local features in the image through convolution operations, while the pooling layer is used to reduce the dimension of image features, reduce the amount of calculation, and enhance the generalization ability of the model. In the VGG19 model, each convolutional layer is followed by a maximum pooling layer, and finally, the extracted features are mapped to specific classification labels through the fully connected layer.

Given an input image $I \in R^{H \times W \times C}$, where H is the height of the image, W is the width of the image, and C is the number of channels of the image (for grayscale images, C=1; for color images, C=3). The process of the convolution operation can be described by the following formula:

$$I^l = K'*I^{l-1} + b^l$$

Among them, $I^l$ represents the output of layer $l$, $K'$ represents the convolution kernel of layer $l$, $b^l$ is the bias term of this layer, and E represents the convolution operation. After each layer of convolution, the size of the feature map changes. Each convolution layer in the VGG19 network uses multiple convolution kernels to extract different features in the image, and the output of each convolution kernel is called a feature map. The convolution operation can effectively extract different levels of features from the input image, including edges, textures, and more complex shapes.

In order to further reduce the dimension of the image and extract important spatial information, the maximum pooling operation is used in VGG19. The purpose of the pooling operation is to reduce the amount of calculation by aggregating local areas while maintaining the important features of the image. The maximum pooling operation is achieved by taking the maximum value in the neighborhood area. The formula for the pooling operation is as follows:

$$I_{pool}^{l} = \max(I^{l}[k,l], \forall (k,l) \in N)$$

Among them, $I_{pool}^{l}$ is the feature map after convolution, $N$ represents the pooled area, and $I_{pool}^{l}$ is the output after the pooling operation. In VGG19, the pooling layer usually performs maximum pooling with a 2x2 window and a stride of 2 to further reduce the size of the feature map.

After completing the convolution and pooling operations, VGG19 uses a fully connected layer to map the extracted features to a high dimension. The fully connected layer is the core part of the neural network. It performs a weighted summation of the output of each layer and applies an activation function for nonlinear transformation. Finally, the output is converted into a probability distribution through the Softmax function to obtain the predicted probability of each category. The formula of the Softmax function is as follows:

$$P(y=i|x) = \frac{\exp(z_i)}{\sum_{j=1}^{N} \exp(z_j)}$$

Among them, $z_i$ is the output of the i-th category, $N$ is the total number of categories, and $P(y=i|x)$ is the probability that sample $x$ belongs to the i-th category. Through the output of the Softmax function, we can get the final classification result and determine whether the image belongs to the pneumonia or non-pneumonia category.

During the training process, VGG19 updates the parameters in the network through the back-propagation algorithm to minimize the loss function [23]. We use the cross-entropy loss function to calculate the classification error of the network. The formula of the loss function is:

$$L = -\sum_{i=1}^{C} y_i \log(p_i)$$

Among them, $y_i$ is the indicator function of the true label, and $p_i$ is the probability of the i-th category output by the model. By minimizing the cross-entropy loss function, VGG19 can continuously optimize the network weights, and finally make the model achieve the best performance in the pneumonia X-ray image classification task.

In general, the VGG19 network effectively extracts local and global features from pneumonia X-ray images through multi-layer convolution, pooling, and full connection operations, and performs final classification predictions through the Softmax classifier. Through back propagation and optimization algorithms [24], VGG19 can continuously improve classification accuracy during training, providing an effective automated solution for early diagnosis of pneumonia.

IV. EXPERIMENT

A. Datasets

The pneumonia X-ray image dataset used in this study comes from the public Kaggle "Chest X-ray Images (Pneumonia)" dataset. This dataset contains about 50,000 chest X-ray images from different patients, covering two categories: normal chest images and pneumonia images. The pneumonia images in the dataset include bacterial pneumonia, viral pneumonia, and other types of pneumonia. The images are all high-resolution anterior-posterior chest X-rays. All images have been annotated, and the annotation information includes normal or pneumonia categories, and whether it is a bacterial or viral infection. This dataset is widely used in the field of medical image analysis, especially pneumonia detection and classification tasks, and provides a valuable foundation for the development of automated diagnostic tools.

Key advantages of this dataset include its large scale, diverse pneumonia cases, and standardized, high-quality grayscale images. During preprocessing, images were scaled, cropped, and normalized to meet VGG19 input requirements. The detailed annotations ensure accurate category matching, enhancing model training quality. This dataset enables the model to learn distinct pneumonia characteristics, improving classification accuracy. The VGG19 model was trained using a learning rate of 0.001 with a batch size of 32 for 100 epochs. The initial learning rate was reduced by 10% every 20 epochs to prevent overshooting.

B. Experimental Results

In order to verify the performance of the proposed VGG19-based pneumonia X-ray image classification model, this study compared four common machine learning and deep learning models, including support vector machine (SVM) [25], XGBoost [26], fully connected multi-layer perceptron (MLP)[27], and ResNet50[28]. As traditional machine learning methods, SVM and XGBoost are usually applied to image classification tasks after feature extraction. They can effectively process high-dimensional data and provide good generalization capabilities. MLP and ResNet50 represent modern deep-learning methods. MLP performs feature mapping through multi-layer neural networks, and ResNet50 uses the structural advantages of residual networks to better deal with the gradient vanishing problem in deep network training. These models provide a diverse comparison benchmark for this study, which helps to comprehensively evaluate the performance of the VGG19 model in pneumonia image classification tasks. The experimental results are shown in Table 1.

Table 1 Experimental Results

| Model | ACC | AUC | F1 | Recall |
|---|---|---|---|---|
| SVM | 0.85 | 0.88 | 0.82 | 0.79 |
| XGBoost | 0.87 | 0.90 | 0.85 | 0.81 |
| MLP | 0.88 | 0.91 | 0.86 | 0.83 |
| ResNet50 | 0.90 | 0.93 | 0.88 | 0.85 |
| VGG19 | 0.92 | 0.95 | 0.90 | 0.87 |

In this experiment, we compared five classification models—SVM, XGBoost, MLP, ResNet50, and VGG19—for pneumonia X-ray image classification. Among these, VGG19 demonstrated the best performance across all evaluation metrics, including accuracy (92%), AUC (0.95), F1 score (0.90), and recall (0.87), highlighting the strengths of deep convolutional neural networks in image classification. ResNet50 also performed well, achieving an accuracy of 90%, AUC of 0.93, F1 score of 0.88, and recall of 0.85. However, its recall and F1 scores were slightly lower than VGG19, indicating a higher likelihood of missing challenging samples. MLP and XGBoost delivered stable but comparatively weaker results. MLP achieved 88% accuracy, 0.91 AUC, 0.86 F1 score, and 0.83 recall, limited by its feature extraction capabilities in complex image tasks. XGBoost, with 87% accuracy, 0.90 AUC, 0.85 F1 score, and 0.81 recall, demonstrated robustness in traditional tasks but lacked optimization for image data. SVM, as a traditional machine learning model, underperformed on this complex image classification task, with 85% accuracy, 0.88 AUC, 0.82 F1 score, and 0.79 recall, largely due to its limited ability to handle high-dimensional data.

Overall, VGG19's superior feature extraction capabilities make it particularly effective for medical image classification, surpassing both traditional machine learning models and other deep learning approaches like ResNet50. The results highlight the advantages of deep learning models, particularly convolutional neural networks, for tasks like pneumonia detection. Future work could explore optimizing VGG19, integrating advanced architectures like self-attention mechanisms, and leveraging techniques such as data augmentation and transfer learning to further enhance performance. Finally, ablation studies on the impact of optimizers, summarized in Table 2, offer additional insights into model optimization.

Table 2 Ablation Experimental Results

| Model | ACC | AUC | F1 | Recall |
|---|---|---|---|---|
| AdaDelta | 0.86 | 0.90 | 0.83 | 0.80 |
| RMSprop | 0.88 | 0.92 | 0.85 | 0.82 |
| SGD | 0.89 | 0.93 | 0.86 | 0.83 |
| Adam | 0.91 | 0.94 | 0.88 | 0.85 |
| AdamW | 0.92 | 0.95 | 0.90 | 0.87 |

Table 2 highlights the performance of various optimizers during model training, with AdamW achieving the best results across all metrics, including accuracy (0.92), AUC (0.95), F1 score (0.90), and recall (0.87), demonstrating its superior optimization and generalization capabilities. In contrast, AdaDelta performed the worst, with accuracy at 0.86 and AUC at 0.90, indicating weaker robustness and higher error rates. The results emphasize the importance of selecting effective optimizers, as adaptive methods like Adam and AdamW improve convergence, reduce overfitting, and enhance model performance, with AdamW's weight decay mechanism being particularly beneficial. AdamW's superior performance across all metrics can be attributed to its adaptive learning rate mechanism combined with weight decay, which prevents overfitting and enhances generalization. In contrast, AdaDelta underperforms due to its reliance on accumulating gradients, which can hinder convergence in complex high-dimensional tasks like pneumonia image classification. RMSprop and SGD perform comparably but lack the regularization benefits of AdamW, explaining their lower recall and F1 scores. Figure 2 further illustrates the training loss trends across different optimizers.

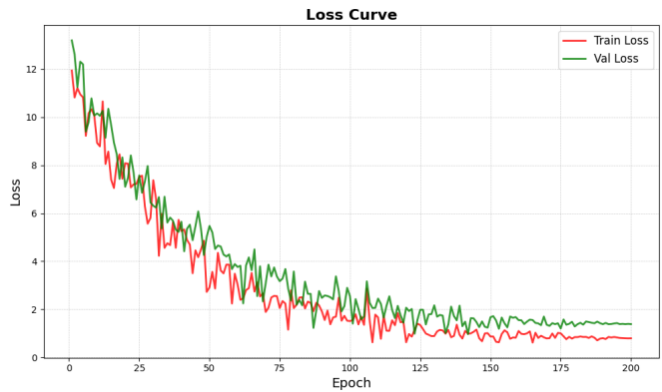

Figure 2 Loss function changes with epoch

This figure shows the change trend of the model's training loss (Train Loss) and validation loss (Val Loss) with the training rounds (Epoch). Overall, both training loss and validation loss decreased significantly with the increase of training rounds, and tended to be stable in the later stage, indicating that the model continued to learn during the training process and gradually approached the optimal state. At the same time, the training loss is always lower than the validation loss, which reflects that the model has stronger fitting ability on the training data but still maintains a certain generalization performance.

Observing from the details of the curve, the verification loss fluctuates greatly in the early stages of training, which may be related to the initialization of model parameters and the high early learning rate. With the deepening of training, the verification loss gradually stabilizes, and the gap with the training loss also remains within a small range, which shows that the model avoids overfitting to a certain extent. However, the late validation loss decreases slower than the training loss, indicating that further improving the model's generalization ability may require additional regularization measures or adjustments to the training strategy.

V. CONCLUSION

This study evaluated the performance of SVM, XGBoost, MLP, ResNet50, and VGG19 for the classification of pneumonia from X-ray images through comparative experiments. The results demonstrate that the VGG19 model excels across multiple evaluation metrics, including accuracy, AUC, F1 score, and recall, significantly outperforming both traditional machine learning models and other deep learning models. Especially in the pneumonia diagnosis task, VGG19 can effectively extract key features in the image and accurately classify them, which helps to improve the efficiency and accuracy of clinical diagnosis.

While models like ResNet50 and MLP also exhibit strong classification performance, VGG19's ability to capture intricate patterns and fine details demonstrates the strength of advanced CNN architectures in tackling complex data, particularly in domains like medical imaging. Traditional machine learning approaches, constrained by their reliance on manual feature engineering, struggle to match the flexibility and power of deep learning frameworks in extracting high-level abstract features critical for tasks such as medical image analysis.

In summary, the VGG19 model's exceptional performance in pneumonia X-ray image classification underscores the transformative potential of deep learning, particularly CNNs, in advancing image analysis tasks. Beyond medical imaging, these findings illustrate how deep learning models can revolutionize automated classification and decision-making in diverse fields, where accurate and efficient feature extraction is paramount. This study reinforces the need for further exploration of CNN-based architectures to tackle broader challenges in high-dimensional data analysis and establishes a strong foundation for developing robust AI-powered diagnostic systems.

References

[1] H. Gupta, N. Bansal, S. Garg, et al., "A hybrid convolutional neural network model to detect COVID-19 and pneumonia using chest X-ray images," *International Journal of Imaging Systems and Technology*, vol. 33, no. 1, pp. 39-52, 2023.

[2] S. Sharma and K. Guleria, "A deep learning model for early prediction of pneumonia using VGG19 and neural networks," in *Mobile Radio Communications and 5G Networks: Proceedings of Third MRCN 2022*, Singapore: Springer Nature Singapore, pp. 597-612, 2023.

[3] S. Phine, "Pneumonia classification using deep learning VGG19 model," Proceedings of the 2023 IEEE Conference on Computer Applications (ICCA), IEEE, pp. 67-71, 2023.

[4] J. Hu, R. Bao, Y. Lin, H. Zhang, and Y. Xiang, "Accurate Medical Named Entity Recognition Through Specialized NLP Models", arXiv preprint, arXiv:2412.08255, 2024.

[5] G. Huang, Z. Xu, Z. Lin, X. Guo, and M. Jiang, "Artificial Intelligence-Driven Risk Assessment and Control in Financial Derivatives: Exploring Deep Learning and Ensemble Models", Transactions on Computational and Scientific Methods, vol. 4, no. 12, 2024.

[6] Y. Yang, C. Tao, and X. Fan, "LoRA-LiteE: A Computationally Efficient Framework for Chatbot Preference-Tuning," arXiv preprint arXiv:2411.09947, 2024.

[7] Z. Qi, J. Chen, S. Wang, B. Liu, H. Zheng, and C. Wang, "Optimizing Multi-Task Learning for Enhanced Performance in Large Language Models", arXiv preprint, arXiv:2412.06249, 2024.

[8] J. Chen, B. Liu, X. Liao, J. Gao, H. Zheng, and Y. Li, "Adaptive Optimization for Enhanced Efficiency in Large-Scale Language Model Training", arXiv preprint, arXiv:2412.04718, 2024.

[9] J. Du, G. Liu, J. Gao, X. Liao, J. Hu, and L. Wu, "Graph Neural Network-Based Entity Extraction and Relationship Reasoning in Complex Knowledge Graphs", arXiv preprint, arXiv:2411.15195, 2024.

[10] Y. Yao, "Self-Supervised Credit Scoring with Masked Autoencoders: Addressing Data Gaps and Noise Robustly", Journal of Computer Technology and Software, vol. 3, no. 8, 2024.

[11] M. Jiang, Z. Xu, and Z. Lin, "Dynamic Risk Control and Asset Allocation Using Q-Learning in Financial Markets", Transactions on Computational and Scientific Methods, vol. 4, no. 12, 2024.

[12] W. Sun, Z. Xu, W. Zhang, K. Ma, Y. Wu, and M. Sun, "Advanced Risk Prediction and Stability Assessment of Banks Using Time Series Transformer Models", arXiv preprint, arXiv:2412.03606, 2024.

[13] Z. Xu, W. Zhang, Y. Sun, and Z. Lin, "Multi-Source Data-Driven LSTM Framework for Enhanced Stock Price Prediction and Volatility Analysis", Journal of Computer Technology and Software, vol. 3, no. 8, 2024.

[14] Z. Zheng, Y. Xiang, Y. Qi, Y. Lin, and H. Zhang, "Fully Convolutional Neural Networks for High-Precision Medical Image Analysis," Transactions on Computational and Scientific Methods, vol. 4, no. 12, 2024.

[15] M. Xiao, Y. Li, X. Yan, M. Gao, and W. Wang, "Convolutional neural network classification of cancer cytopathology images: taking breast cancer as an example," Proceedings of the 2024 7th International Conference on Machine Vision and Applications, pp. 145–149, Singapore, Singapore, 2024.

[16] W. Wang, Y. Li, X. Yan, M. Xiao, and M. Gao, "Breast cancer image classification method based on deep transfer learning," Proceedings of the International Conference on Image Processing, Machine Learning and Pattern Recognition, pp. 190–197, 2024.

[17] S. Lu, Z. Liu, T. Liu, and W. Zhou, "Scaling-up medical vision-and-language representation learning with federated learning," Engineering Applications of Artificial Intelligence, vol. 126, p. 107037, 2023.

[18] A. Shen, M. Dai, J. Hu, Y. Liang, S. Wang, and J. Du, "Leveraging Semi-Supervised Learning to Enhance Data Mining for Image Classification under Limited Labeled Data," arXiv preprint, arXiv:2411.18622, 2024.

[19] Y. Xiao, "Self-Supervised Learning in Deep Networks: A Pathway to Robust Few-Shot Classification," arXiv preprint, arXiv:2411.12151, 2024.

[20] C. Ruan, C. Huang, and Y. Yang, "Comprehensive Evaluation of Multimodal AI Models in Medical Imaging Diagnosis: From Data Augmentation to Preference-Based Comparison," arXiv preprint, arXiv:2412.05536, 2024.

[21] X. Yan, W. Wang, M. Xiao, Y. Li, and M. Gao, "Survival prediction across diverse cancer types using neural networks," Proceedings of the 2024 7th International Conference on Machine Vision and Applications, pp. 134–138, 2024.

[22] Q. Sun, T. Zhang, S. Gao, L. Yang, and F. Shao, "Optimizing Gesture Recognition for Seamless UI Interaction Using Convolutional Neural Networks," arXiv preprint, arXiv:2411.15598, 2024.

[23] Z. Liu and J. Song, "Comparison of Tree-Based Feature Selection Algorithms on Biological Omics Dataset," Proceedings of the 5th International Conference on Advances in Artificial Intelligence, pp. 165-169, November 2021.

[24] S. A. Aljawarneh and R. Al-Quraan, "Pneumonia detection using enhanced convolutional neural network model on chest X-ray images," Big Data, 2023.

[25] M. Mardianto, A. Yoani, S. Soewignjo, et al., "Classification of pneumonia from chest X-ray images using support vector machine and convolutional neural network," *International Journal of Advanced Computer Science & Applications*, vol. 15, no. 6, 2024.

[26] M. El-Ghandour and M. I. Obayya, "Pneumonia detection in chest X-ray images using an optimized ensemble with XGBoost classifier," *Multimedia Tools and Applications*, pp. 1-31, 2024.

[27] A. Çelik and S. Demirel, "Enhanced pneumonia diagnosis using chest X-ray image features and multilayer perceptron and k-NN machine learning algorithms," *Traitement du Signal*, vol. 40, no. 3, p. 1015, 2023.

[28] S. A. Rachman, D. C. Bagaskara, R. Magdalena, et al., "Classification of pneumonia based on X-ray images with ResNet-50 architecture," Proceedings of the 3rd International Conference on Electronics, Biomedical Engineering, and Health Informatics: ICEBEHI, 2022.